# J-Score: A Robust Measure of Clustering Accuracy

Navid Ahmadinejad[1], Li Liu[1,2,3]*

[1] College of Health Solutions, Arizona State University, Phoenix, AZ, 85004, USA
[2] Biodesign Institute, Arizona State University, Phoenix, AZ, 85281, USA
[3] Department of Neurology, Mayo Clinic, Scottsdale, AZ 85259, USA

* Corresponding Author:
Li Liu (liliu@asu.edu)

# Abstract

**Background.** Clustering analysis discovers hidden structures in a data set by partitioning them into disjoint clusters. Robust accuracy measures that evaluate the goodness of clustering results are critical for algorithm development and model diagnosis. Common problems of current clustering accuracy measures include overlooking unmatched clusters, biases towards excessive clusters, unstable baselines, and difficult interpretation. In this study, we presented a novel accuracy measure, J-score, that addresses these issues.

**Methods.** Given a data set with known class labels, J-score quantifies how well the hypothetical clusters produced by clustering analysis recover the true classes. It starts with bidirectional set matching to identify the correspondence between true classes and hypothetical clusters based on Jaccard index. It then computes two weighted sums of Jaccard indices measuring the reconciliation from classes to clusters and vice versa. The final J-score is the harmonic mean of the two weighted sums.

**Results.** Via simulation studies, we evaluated the performance of J-score and compared with existing measures. Our results show that J-score is effective in distinguishing partition structures that differ only by unmatched clusters, rewarding correct inference of class numbers, addressing biases towards excessive clusters, and having a relatively stable baseline. The simplicity of its calculation makes the interpretation straightforward. It is a valuable tool complementary to other accuracy measures. We released an R/jScore package implementing the algorithm.

# Introduction

Cluster analysis is an unsupervised data mining technique that partitions data into groups based on similarity [1]. It is a valuable approach to discover hidden structures and has broad application in pattern recognition. Many clustering methods have been developed and data sets are subject to cluster analysis constantly [2-4]. To evaluate algorithm performance, select models, and interpret partition structures, a robust measure of clustering accuracy is imperative.

Cluster analysis speculates that subsets of input data belong to different classes and aims to discover these classes by partitioning data into hypothetical clusters. When true class labels of input data are known, accuracy of clustering results can be assessed on how well hypothetical cluster assignments recover true class labels [5]. Intuitively, the assessment involves first establishing the correspondence between true classes and hypothetical clusters (i.e., set matching) [6], then quantifying the overall goodness of match. For example, given a true class, the hypothetical cluster sharing the largest number of data points with it can be regarded as the best match. The fraction of total unmatched data points aggregated over all classes is then reported as an H-score [7]. The cluster best matched to a class can also be determined to maximize the harmonic mean of precision and recall rates (i.e., F1-score). Weighting F1-scores over all classes produces an F-score to represent the overall accuracy [8]. Because set matching reports only best clusters matched to each class, unmatched clusters do not contribute to the overall accuracy score. When two hypothetical partition structures differ only by unmatched clusters, these accuracy measures are unable to distinguish them [9, 10].

To address this "problem of matching", several measures have been developed that circumvents set matching. Instead of linking specific classes to clusters, these measures quantify mutual agreement between all classes and all clusters. For example, to compare a true partition structure and a hypothetical partition structure of the same data set, Rand index RI [11] and its adjusted form ARI [12] count pairs of data points that are consistently clustered together or separately. Variation of information (VI) criterion and its normalized form NVI compute the amount of information lost and gained in changing from one partition structure to another partition structure. V-measure combines homogeneity and completeness of clusters [10], which is equivalent to normalized mutual information (NMI) [13]. However, most of these measures are information theoretic based and are biased towards excessive small-sized clusters [14-16]. Furthermore, circumventing set matching sacrifices the ability to discover correspondence between classes and clusters, which is the primary goal of clustering analysis. Thus set-matching-free measures have limited utility in model diagnosis. Although it is possible to perform post hoc set matching or score adjustment, validity and interpretability of the outputs do not align to the original scores.

In this study, we introduce J-score, a novel clustering accuracy measure that supports four desirable properties. First, it performs set matching to identify correspondence between classes and clusters. Second, set matching is bidirectional, which finds clusters best aligned to classes and vice versa. Subsequently accuracy calculation incorporates both matched and unmatched clusters to address the "problem of matching". Third, it is based on Jaccard index, a non-entropy-based metrics, which we show mitigates excessive number of clusters. Fourth, its value is bounded between 0 and 1. We illustrate behavior of *J*-score with extensive simulations.

## Materials & Methods

<u>J-Score</u>

*Bidirectional set matching*: Suppose that a data set contains $N$ data points belonging to $T$ true classes, and cluster analysis produces $K$ hypothetical clusters. To establish the correspondence between $T$ and $K$, we first consider each class as reference and identify its best matched cluster ($T \to K$). Specifically, for a class $t \in T$, we search for a cluster $k \in K$ that has the highest Jaccard index,

$$I_t = \max_{k \in K} \frac{|V_t \cap V_k|}{|V_t \cup V_k|}$$

where $V_t$ and $V_k$ are the set of data points belonging to class $t$ and cluster $k$, respectively, and $|\cdot|$ denote the size of a set. We then consider each cluster as reference and identify its best matched class ($K \to T$) using a similar procedure. For a cluster $k \in K$, we search for a class $t \in T$ with the highest Jaccard index,

$$I_k = \max_{t \epsilon T} \frac{|V_t \cap V_k|}{|V_t \cup V_k|}$$

*Calculating overall accuracy*: To quantify the overall accuracy, we aggregate Jaccard indices of individual clusters and classes, accounting for their relative sizes (i.e., number of data points). We first compute a weighted sum of $I_t$ across all classes as $R = \sum_{t \epsilon T}(\frac{|V_t|}{N} I_t)$, and a weighted sum of $I_k$ across all clusters as $P = \sum_{k \epsilon K}(\frac{|V_k|}{N} I_k)$. We then take their harmonic mean as $J$ score,

$$J = \frac{2 \times R \times P}{R + P}$$

The value of a $J$ score is bounded between 0 and 1. The minimum value 0 is derived when there is no overlap between hypothetical clusters and truth clusters. The maximum value 1 is derived when hypothetical clusters match true classes perfectly.

*Implementation*: We implemented *J*-score calculation in R language and released an jScore package in R/CRAN repository.

Simulations

To simulate an input data set $D$ with known class labels, we generated random numbers based on Gaussian distributions $G(\mu, 0.05)$ where $\mu$ is the mean and 0.05 is the fixed standard deviation. Data points generated from the same Gaussian distribution belonged to the same class. We then mixed data points of different classes to produce the input data. Class labels of these data points were ground truth.

Given an input data set $D$ with $N$ data points, we used three approaches to simulate a hypothetical partition structure. The first approach simulated a pre-determined partition structure, in which the total number of clusters $K$, the size of each cluster $N_k$, and the assignment of each data point to a cluster were specified manually. The second approach simulated a random partition structure. Here, only the value of $K$ was pre-specified. $N_k$ was determined by randomly choosing $K$ integers that summed to $N = \sum_1^K N_k$. Assignment of data points to clusters was also random, which was achieved by first repeating each $k$ value by $N_k$ times to create an ordered list of cluster labels, then permutating these labels.

The third approach simulated splitting or merging classes. Given the pre-specified value of $K$, classes to be split or merged and the splitting ratio were randomly selected under the constraint that $N = \sum_1^K N_k$.

Computation and comparison of various accuracy measures

We compared J-score with commonly used clustering accuracy measures. To compute NMI and ARI scores, we used the R/aricode package. To compute V-measure and F-score, we used the Python scikit-learn package. To compute F-score and H-score, we used an in-house developed R functions based on the published algorithms [7, 17].

## Results

J-Score addresses the "problem of matching".

Given a data set, the number of hypothetical clusters may be equal to, less than, or greater than the number of true classes. In all three scenarios, our simulations showed that unidirectional $T \rightarrow K$ matching could lead to the "problem of matching", and J-score rectified the biases via bidirectional matching.

For ground truth, we generated 100 random numbers $D_i$ ($i = 1, \dots, 100$) belonging to three classes. Specifically, $D_{1,\dots,10}$ from a Gaussian distribution $G(1, 0.05)$ constituted class $T_a$, $D_{11,\dots,40} \in G(2, 0.05)$ constituted class $T_b$, and $D_{41,\dots,100} \in G(3, 0.05)$ constituted class $T_c$ (**Fig. 1A**).

We first examined hypothetical partition structures that contained more clusters than classes. In one simulation, we grouped the data points into four clusters (**Fig. 1B**). Cluster $K_1$ consisted of all data points from class $T_a$; Cluster $K_2$ consisted of all data points from class $T_b$; Cluster $K_3$ consisted of $D_{41,\dots,80}$ that are two thirds of data points from $T_c$; and cluster $K_4$ consisted of the remaining data points from $T_c$. We assessed the accuracy of this hypothetical partition structure using H-score, F-score and J-score. Because the number of hypothetical clusters exceeded the number of true classes, unmatched clusters were inevitable. Indeed, despite different $T \rightarrow K$ algorithms, all three approaches identified $K_1$, $K_2$, and $K_3$ as the cluster best matched to $T_a$, $T_b$, and $T_c$, respectively, and left $K_2$ unmatched. The H-score (0.20) and the F-score (0.88) were calculated based on the goodness of $T_a \to K_1$, $T_b \to K_2$, and $T_c \to K_3$ matches, with no information contributed by $K_4$. The "problem of matching" occurred when we split the unmatched $K_4$ cluster into two clusters $K_{4.1}$ and $K_{4.2}$ (**Fig. 1C**). This splitting obviously reduced the overall accuracy of the partition structure. But the H-score and the F-score remained unchanged because $K_{4.1}$ and $K_{4.2}$ were unmatched and did not contribute to the final scores. J-score solved this problem via bidirectional matching. The unmatched clusters in the $T \rightarrow K$ matching step were rescued in the $K \rightarrow T$ matching step. Then, all clusters and classes contributed information to the final J-score. As expected, the J-score dropped from 0.77 to 0.75 after the splitting.

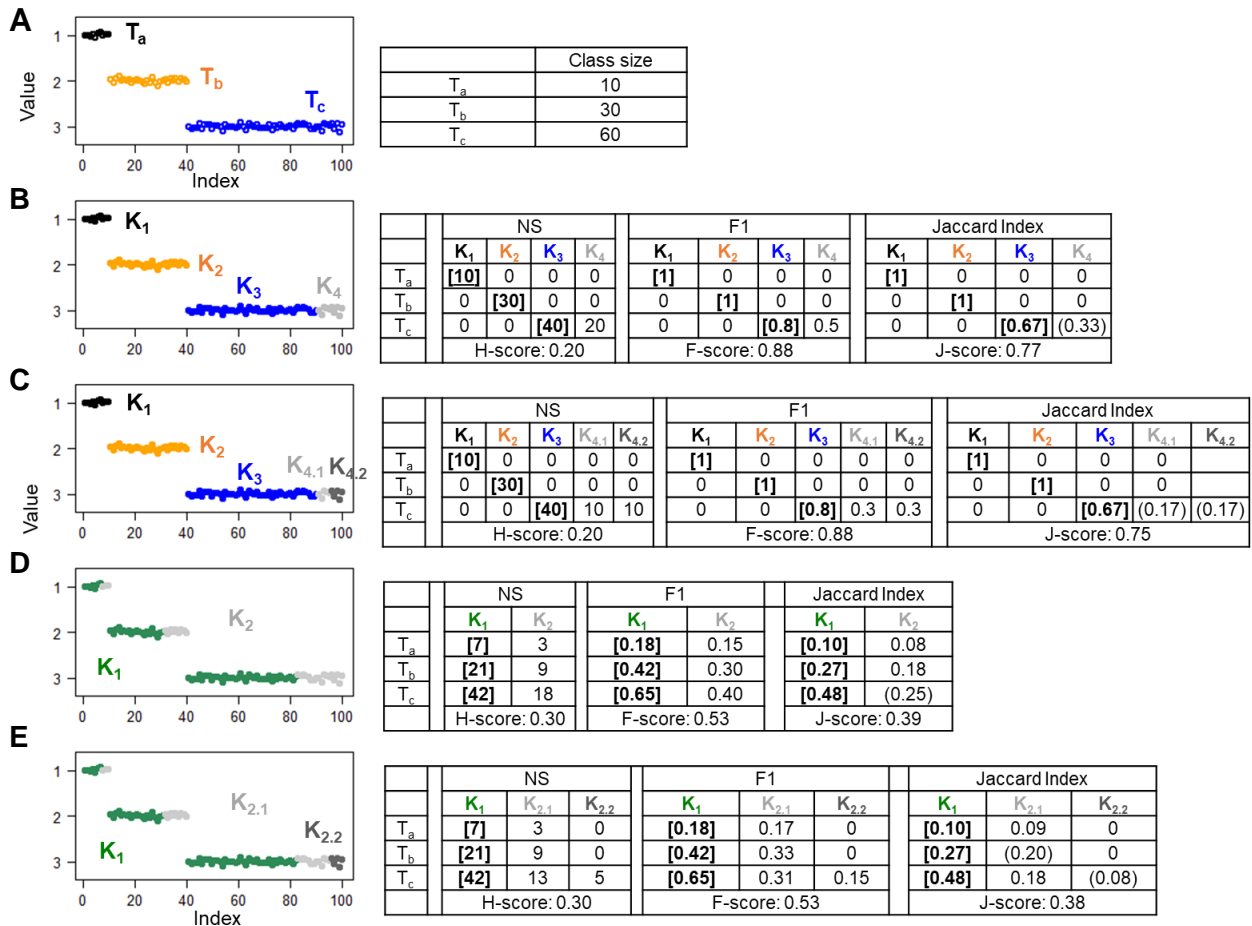

**Figure 1. Simulations to illustrate the "problem of matching".** Scatterplots show 100 data points with indices ranging from 1 to 100 and values randomly generated from Gaussian distributions. Colors denote different classes or clusters. (**A**) The ground truth partition structure contains 3 classes. Class sizes are displayed in the table. (**B-E**) Hypothetical partition structures. Various scores and set matching results are displayed in the tables next the scatterplots. H-score and F-score perform unidirectional $T \to K$ matching based on number of shared data points (NS) and F1 metric, respectively. Best matched clusters are enclosed in square brackets. J-score performs bidirectional matching. Best matched classes are enclosed in parentheses.

Unmatched clusters exist even when there are fewer clusters than classes. To illustrate the advantage of J-score in these scenarios, we simulated a hypothetical partition structure that grouped the data points into two clusters. Specifically, cluster $K_1$ mixed 70% of the data points from each class. The remaining data points were grouped into cluster $K_2$ (**Fig. 1D**). After the $T \rightarrow K$ matching step, $K_1$ was repeatedly identified as the best matched cluster for all three classes, and $K_2$ was unmatched. In another hypothetical partition structure, we kept cluster $K_1$ untouched and split $K_2$ into $K_{2.1}$ and $K_{2.2}$ that were unmatched as well (**Fig. 1E**). Again, because these two hypothetical partition structures differ only by unmatched clusters, H-score and J-score failed to distinguish them. J-score correctly reported a higher value for the first structure than the second structure (0.39 vs. 0.38).

J-score reflects correct inference of class numbers and addresses biases towards excessive clusters.

An important objective of cluster analysis is to infer the number of classes in input data. A robust accuracy measure shall reward a hypothetical partition structure if it contains the correct number of clusters and penalize incorrect ones proportionally. Because assignment of individual data points to each cluster also influences the accuracy, we used simulated hypothetical partitions structures that split or merge true classes to minimize the confounding effect. For ground truth, we generated 1000 random numbers belonging to 10 classes. We varied the number of clusters $K$ from 1 to 50. For each hypothetical partition structure, we computed seven accuracy measures including J-score, F-score, V-measure, RI, ARI, NMI, and NVI. We repeated this process 200 times for each $K$ value and examined the mean and variance of each score. Since the number of true classes was 10, we expected these scores should peak at $K = 10$. This was indeed the case for J-score, F-score, ARI, and NMI (**Fig. 2A**). These scores also decreased sharply as $K$ deviated from 10, penalizing both deficient and excessive clusters. However, V-measure, RI, and NVI peaked incorrectly at $K = 13$, overestimating the number of classes (**Fig. 2B**). These two measures penalized excessive clusters only slightly. Even when $K = 50$ that was an overestimation by 400%, these scores decreased only by 5% from their peak values.

To examine if J-score remains robust when the number of classes decreases and the class size increases, we simulated 1000 random numbers belonging to 5 classes. Again, J-score peaked at correct inferences of class counts (**Fig. 2C**) while all the other scores overestimated (**Fig. 2D**). These results are consistent with previous reports of biases towards excessive clusters using existing accuracy measures [14-16]. J-score is vigorous in this perspective.

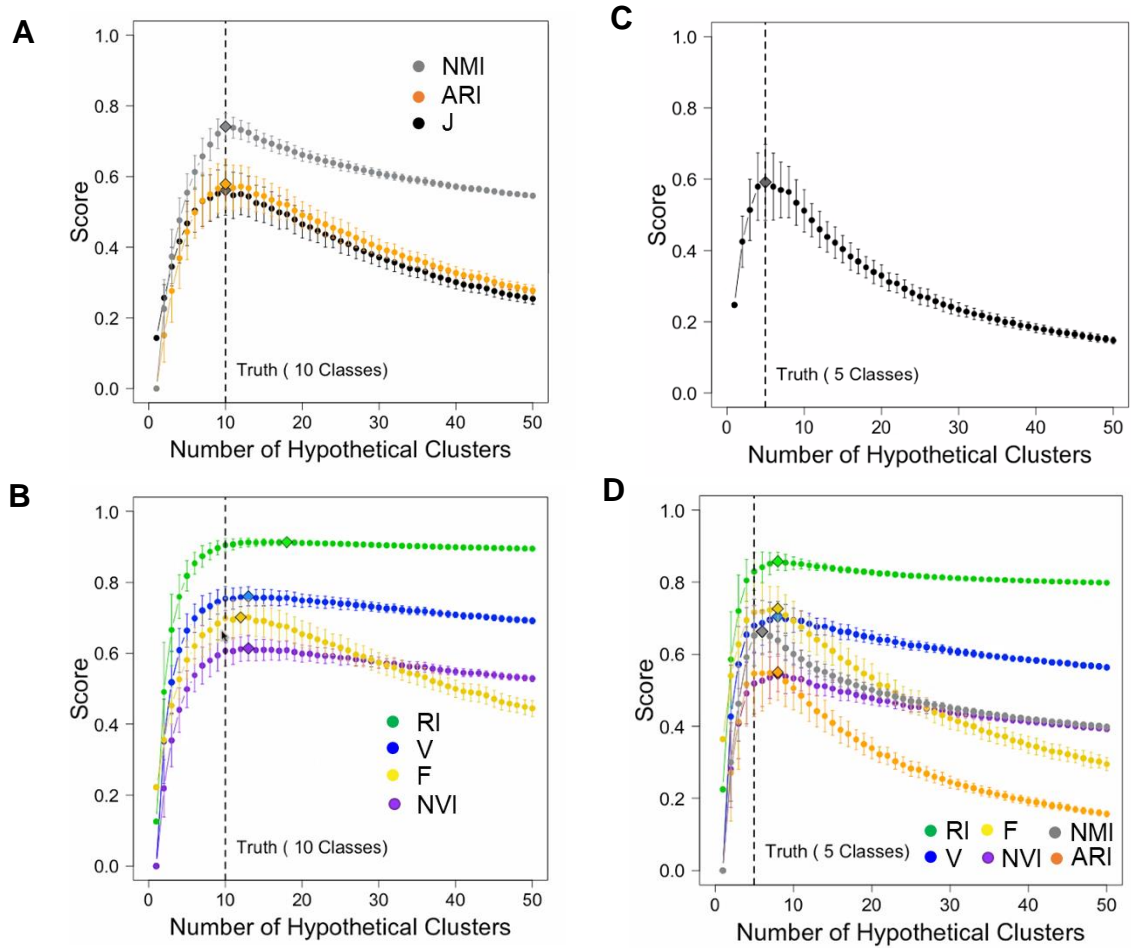

**Figure 2. Simulations to illustrate inferences of the number of classes. (A, B)** Simulations using a data set containing 1,000 data points from 10 true classes. For each accuracy measure, mean scores of 200 hypothetical partition structures containing a given number of clusters are displayed. Error bars represent standard deviations. Diamonds mark the inferred number of classes by the corresponding accuracy measures. Some measures including J-score made correct inferences (A) and others made incorrect inferences (B). (**C, D**) Simulations using a data set containing 1,000 data points from 5 true classes. Only J-score made the correct inference (C) and others made incorrect inferences (D).

### J-score has a relatively stable baseline.

When using an accuracy measure to evaluate multiple hypothetical partition structures, relative accuracy values are sufficient to identify which one best fits the ground truth, and a universal baseline is not needed. However, a stable baseline facilitates interpretation of a single accuracy value [16]. To examine the baselines of J-score and other measures, we computed similarities between random hypothetical partition structures. Specifically, we simulated 1,000 random numbers, randomly assigned them to $K$ clusters, and varied $K$ values from 2 to 50. For each $K$, we generated two random hypothetical partition structures and measured their similarities using various measures. We repeated this process 50 times for each $K$ value. Because these partition structures were random, their mean pairwise similarities shall be close to the lowest value of the corresponding accuracy measure and stay constant regardless of the $K$ value. This was indeed the case for ARI that is designed deliberately to maintain a stable baseline at 0 (**Fig. 3**). The second best was J-score, whose baseline was stabilized around 0.08 after the $K$ value reached 12 and higher, i.e., the mean cluster size was less than 83. However, when K was small, many data points were grouped together simply by chance, leading to high similarities between random partition structures and consequently high baseline values. F-score showed a similar pattern as J-score, but with significantly higher baseline values. The remaining measures, RI, V-measure, NVI, and NMI had exponentially increasing baselines across the entire range of K values we tested with no signs of stabilization.

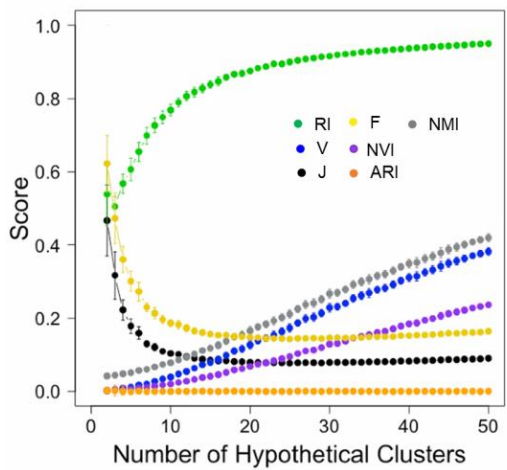

**Figure 3. Simulations to illustrate the baselines of various measures.** For each K, the mean score of 200 pairwise similarities between random hypothetical partition structures are displayed. Error bars represent standard deviations.

## Discussions

Clustering accuracy can be estimated with or without knowing ground truth class labels, which affects the application and interpretation of an accuracy measure. Ground truths are mostly available in simulated or curated data used for algorithm development and evaluation. In these applications, besides a single score to estimate the overall goodness, information about specific clusters or classes is also valuable for diagnosis. We developed J-score that supports these functionalities and addresses common issues including the "problem of matching" and biases towards excessive clusters.

The advantages of J-score over H-score and F-score measures are achieved via bidirectional matching to recover unmatched clusters. Although we did not compare other set matching algorithms such as those using distances between centroids to assign clusters to classes, the "problem of matching" applies because they are also based on unidirectional matching. To promote the adoption of bidirectional set matching, we included a utility function in our R/jScore package that takes in a table of arbitrary pairwise cluster/class similarity scores, performs bidirectional set matching, and returns the correspondences. In fact, because Jaccard index and F1-score are monotonically related, one can modify F-score using this utility function and potentially solve the "problem of matching" as well.

Compared to measures based on mutual agreement, J-score have three advantages. First, J-score identifies correspondence between classes and clusters and quantifies pairwise similarities, which is valuable information for model diagnosis and result interpretation. One can argue that post hoc set matching may be a remedy for mutual agreement-based measures. However, it does not fix the disconnect between matched sets and final accuracy scores, which can mislead model diagnosis and evaluation.

Second, J-score does not suffer from biases toward excessive clusters. A partition structures with many small clusters risk overfitting. We show that J-score, without additional normalization or correction, performs better than or similarly as ARI, VI and NMI measures that involves complicated computations. Third, J-score has a relatively stable baseline when the average cluster size is not too big, making interpretation of a single accuracy value straightforward. Many existing measures lack this property except for ARI that specifically adjusts for chance [16]. However, J-score achieves this without special adjustment, keeping the algorithm simple and intuitive.

However, ground truth is not available in real-world applications. In these cases, clustering accuracy can be estimated using internal validity measures such as the R2 score, the silhouette index, and the SDbw index that quantify within-cluster similarity and between-cluster similarity [18]. Alternatively, one can apply two different clustering algorithms to the same data set and interpret reproducibility as accuracy. This motivated the development of several mutual agreement-based measures that are later repurposed for comparing hypothetical and ground true partition structures. The VI and NMI measures are among this category. Because J-score performs bidirectional set matching, it is theoretically feasible to use it to assess reproducibility or derive consensus clusters. However, this is beyond the scope of this manuscript and we plan to evaluate its performance in this aspect in a separate study.

There are no perfect clustering accuracy measures. Like other available measures, J-score has its weaknesses. For example, it is not a true metric because it does not satisfy the triangular inequality. However, J-score has many desirable and complementary properties to existing measures, making it a valuable addition to the toolbox.

## Conclusions

J-score is a simply and robust measure of clustering accuracy. It addresses the problem of matching and reduces the risk of overfitting that challenge existing accuracy measures. It will facilitate the evaluation of clustering algorithms and clustering analysis results that are indispensable in big data analytics.